# Rescue Robotics in Bore well Environment

Manish Raj, P.Chakraborty and G.C.Nandi
Indian Institute of Information Technology, Allahabad
{ rajmanish.03, pavan, gcnandi}@gmail.com

*Abstract*— **A technique for rescue task in bore well environment has been proposed. India is facing a distressed cruel situation where in the previous years a number of child deaths have been reported falling in the bore well. As the diameter of the bore well is quiet narrow for any adult person and the lights goes dark inside it, the rescue task in those situations is a challenging task. Here we are proposing a robotic system which will attach a harness to the child using pneumatic arms for picking up. A teleconferencing system will also be attached to the robot for communicating with the child.**

## I. INTRODUCTION

In India, recently we have witnessed some of the most tragic but helpless incidents which touched us deeply and forced us to look after the matter seriously. As the statistics suggests in the consecutive years starting from 2006, still more than 30 deaths occurred while stuck in bore well [1]. The most mournful fact in that figure is that 92% of that victim is under the age of 10. The children were playing around the bore well unaware of the fact that the bore well was waiting for them in the form of a death trap. After slipping in the rotten congested pitch black environment they were waiting for the help to come. But the lack of oxygen and deathly atmosphere has taken their life slowly before the rescue team can reach them.

The incident of losing lives trapped in bore-well was highlighted in 2006 where a 5 year old child named Prince [4] was rescued by Indian Army experts after a tough combat which lasted 49 hours. The boy showed tremendous survival instinct by remaining calm and being co-operative with survivors. Statistics reveal [2] [3] that not many kids were as lucky as Prince, many of them died, some received public attention, while many went unnoticed. Another incident in Indore took place in the same year where a child name Deepak stuck in the pit hole and died for the lack of oxygen. We have tried to summarize the incidents in this concern.

It's our agony that on April 7,2007 in Village Adsar in

Bikaner district (Rajasthan),we witnessed the death of a two year-old girl named Sarika who had fallen in **a 155-feet** deep open bore-well and on the same day, a two-year-old girl, Kinjal Man Singh Chauhan, fell in an open bore-well in village Madeli (Gujarat) and died. On February 6, 2007, a two-year-old boy, Amit, fell in a 56-feet deep well in a village near Katni (MP) and died. On March 9, 2007, in Karmadia (Gujarat) three year-old died due to same. On June 17,2007 an open bore-well in village Shiroor (Pune, Maharashtra) claimed the life of a five-year-old child. Six-year old Suraj lost his life when he fell in a 180-feet-deep bore-well in village Nimada (Jaipur, Rajasthan) on July $4^{th}$,2007. On August 4,2007 six-year-old Kartik died when he slipped in a 200-feet-deep open bore-well in village Botala Gudur (Andhra Pradesh). This was the year of sorrow as small accidents were taking the lives of innocent children. The most common thing in those incidents was a fact of lack of technology. This didn't stopped hear yet.

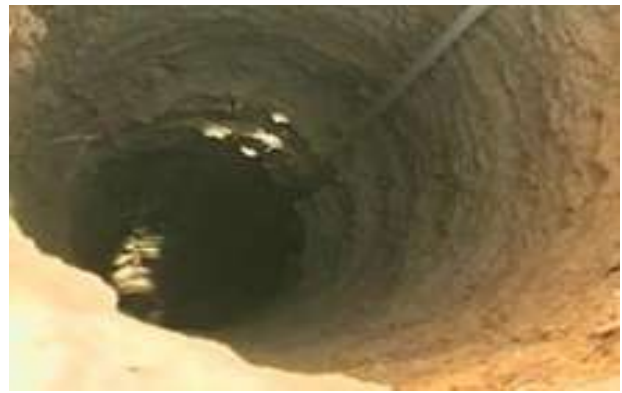

Fig. 1. Image of a bore-well .Image taken during the rescue operation[2]

On March 25, 2008 a three-year-old girl, Vandana, fell in a 160-feet-deep open bore-well in village Tehra near Agra. 2-year old Sonu fell in 150 feet deep bore well pit in the northern state of Uttar Pradesh. He was brought out dead after four days of rescue operation. In 2009, Kirtan Pranami, an 11-year-old boy from Palanpur in Gujarat died after he fell into a 100ft (30m) bore-well. Within months, two-year-old Darawath Mahesh fell into a 35ft (10m) bore-well in Warangal in Andhra Pradesh and died. Five-year-old child who fell into a 250-feet deep bore-well in Jaipur in 2009 was also saved. Four-year-old Anju Gujjar was rescued also from a 50-foot deep open bore-well in Rajasthan.

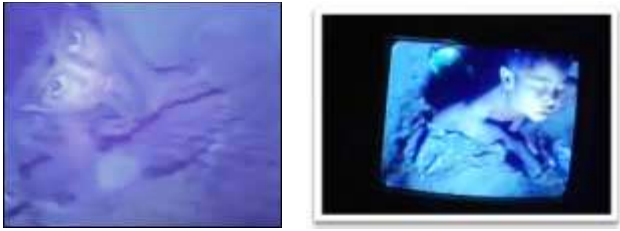

Fig. 2 Image of the boy named Prince felt in the bore-well. Image Courtesy Zee News[2][3]

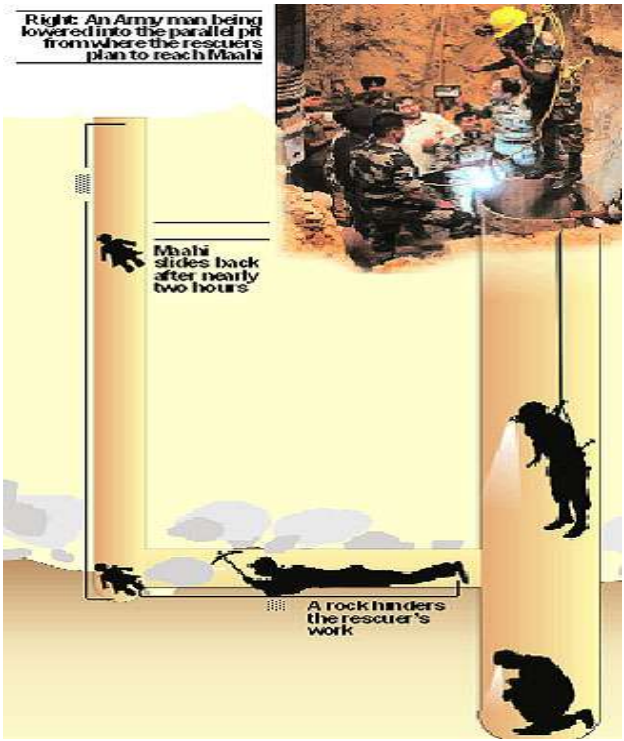

Fig. 3 .Graphical representation of the rescue task by the Indian Army Jawans[4]

The redemption of 4-year old Mahi (2012) took army experts 86 hours ordeal combat which led to the death of the poor kid. Other sad incidents in 2012 was the deadly incidents of 4-year old boy of Tamil Nadu (2011), 1-year old Payal of Indore (2012), 12-year old Bakul of Gujarat (2012), or 17-year old Roshan of Howrah, West Bengal (2012).

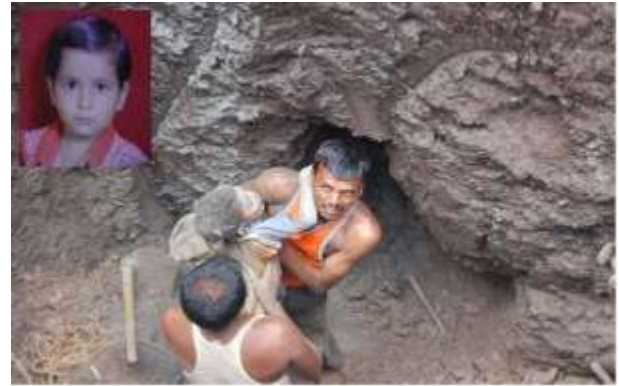

Fig. 4. The child (Mahi) found dead after 86 hours of operation. Image taken during the rescue task[5]

The sadness caries out even now after 6 years from the first case that gain huge limelight and support from the media. Each time something happens we find ourselves ill-equipped to deal with the crisis and the precious time elapses.

We first observed an extensive approach [5] from the Indian Army military jawans (L&T ,GMR) in the case of Mahi (2012). As the rescuer team found that the task of picking up the kid in a straight path is not possible, they started to dig up another well in form of a well in not so far distance from the accident spot. An army man was been lowered into the new parallel pit where he started to dig a vertical lane to reach Mahi. The rescue operation is graphically illustrated in the Fig. 3. The operation lasted for around 86 hours and at last what her parents got was the body without internal soul.

II. PROPOSED SYSTEM DESCRIPTION

Our proposed system consists of mainly two round plates. The full hardware system has been illustrated by the Fig.5. A mechanical system will be attached to the higher plate which will try to release two linear actuation units which will hold the robot in position by pushing the wall of the bore-well. Another mechanical gear system will be attached which will rotate the lower plate to get position it in plane with the victim. Two arms will be attached to the lower plate. Two high resolution cameras will be attached downwards in the lower position of the lower plate. The high resolution cameras will provide the view of the well environment which will be highly helpful in teleoperating the two arms[6][7]. As the bore well environment is a dark environment the robot will be having lights which will provide enough lighting conditions for the operation of the robot. The pneumatic arms will be having another two individual cameras for each arm which will publish the view of the arms. A chest mount harness will be attached with the robot which will be highly essential in picking up the victim from the bore well. The victim can see and communicate with their family members through teleconferencing by the communication system which consists of a LCD screen with high definition audio systems[8][9].

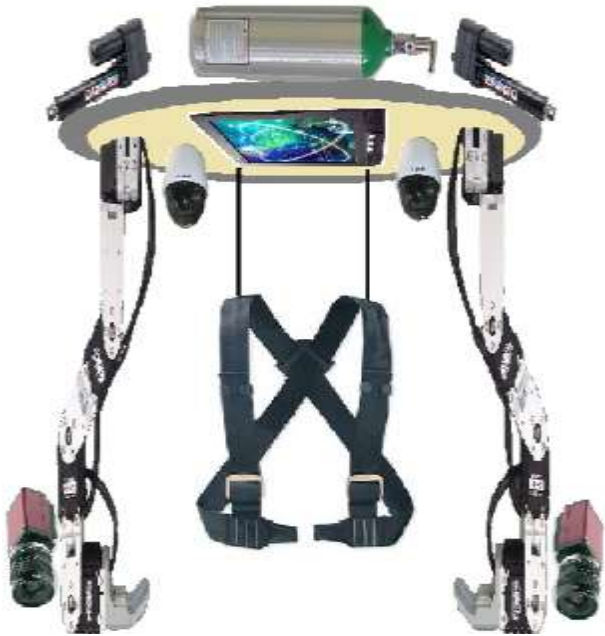

Fig.5. Illustrated Hardware module

Oxygen supply will be provided by oxygen pipes which will go with the robotic system. There will also be an supplementary oxygen mask which can be used by the robotic arms in emergency situations

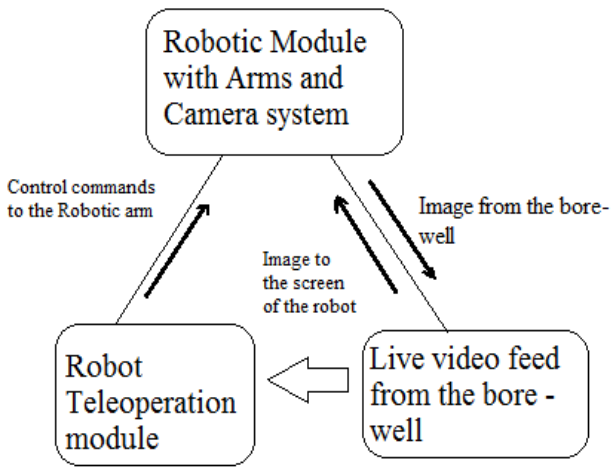

Fig. 6 Data flow diagram of the system

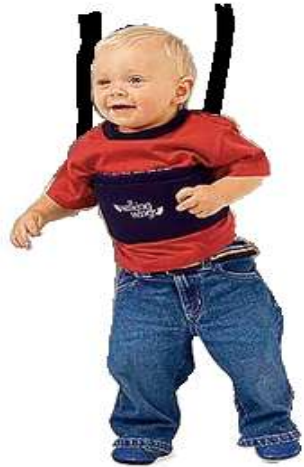

Fig.7. Image of a boy strapped with a chest mount harness

The proposed rescue task is been presented by the illustration in the Fig.8 .As first the robot will be implemented hanging inside the bore-well with the help of ropes. After reaching a certain distance from the victim the robot will be stopped for deciding the pick-up position as we can see the whole view of the bore-well through images being published by the high resolution cameras[10][11]. As we can observe the whole situation and control the two arms of the robot, the complex task of deciding the pick-up points and getting the victim inside the harness will get simplified without the risk of getting any military jawan or concerned person inside the bore-well. The linear actuators will fix the robot in the exact pick up position by pressing the walls of bore-well. The robotic arms will be teleoperated to attach the straps of harness around the victim. As from the upper mounted cameras we can only observe the upper view of the situation, the task will get complicated if the victim will not be co-operative or get senseless. We will be providing another two cameras situated at the tip of the arms which will provide the view in-front of the arms and the arms could be smoothly teleoperated. If the victim will be in a panic mode by less oxygen condition, the robot will also provide oxygen supplies through the oxygen supply pipelines as well as from the back up oxygen cylinder. As the robot attaches the harness to the victim, the victim would be ready to pickup. As various commercial harness are available today we choose to use the chest harness to reduce the complexity of the task. A chest harness is capable of lifting a 15 yr old boy easily. The chest harness lifting a boy is shown in Fig.7[12][13].

The whole scenario will be feeded live through the communication module which will publish the images from the cameras of the robot. The family members of the victim can also see the condition where their family member is been stuck. Some of them can also collaboratively help the rescue process by soothing the victim with their affectionate talks. The data flow diagram of the communication system has been presented in the Fig.6. The victim can also communicate with their family members through the teleconferencing system. The person operating the robotic arms can also the view the images from the live cameras at the top, as well as the cameras situated at the tip of the arms. The arm-tip cameras will provide the view of the route of the arms for attaching the straps of the harness, also bringing the food-bag or oxygen mask to the victim.

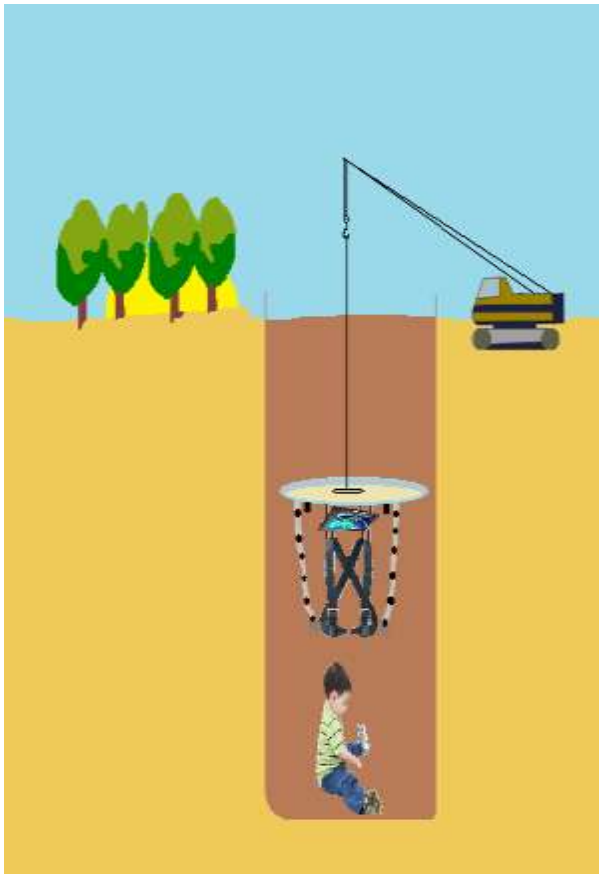

Fig.8 Illustrated image of the proposed rescue task

## IV. CONCLUSION

A robotic framework for rescue robotics in bore-well environment has been proposed here. Deeply observing those incidents and looking at the current circumstances we feel that we need to develop such framework for saving those innocent lives. In addition there is a whole new research area waiting ahead us which deals with lots of challenges relating to mapping in unknown environment, real-time teleoperation in low lighting conditions, arm manipulation system. Rather than the technical development we would be highly satisfied if it can fulfill the most important aspect of the project, which is to *save a life*.

ACKNOWLEDGEMENT

The authors would like to present their sincere grievance to the families who lost their dear ones in those situations. We present our standing ovation to those military and non-military persons who tried their best to save lives. We give special thanks to Mrs.Debashree Makhal who encouraged us and showed us the path for this humanitarian effort. We thank the news media namely Zee News, The Hindu, Times of India, NDTV and the others .Our grievance to those children who lost their lives in that incidents.

Deepak of Indore (2006), 2-year old Kinjal Man Singh Chauhan of Gujarat (2007), 2-year old Amit of Madhya Pradesh (2007, 11-year old Kirtan Pranami of Palanpur, Gujarat (2009), D. Mahesh of Warangal (2010), 4-year old boy of Tamil Nadu (2011), 1-year old Payal of Indore (2012), 12-year old Bakul of Gujarat (2012), 4-year old Mahi of Manesar (2012) or 17-year old Roshan of Howrah, West Bengal (2012).